\documentclass[10pt,twocolumn,letterpaper]{article}

\usepackage[review]{cvpr}      

\usepackage{graphicx}
\usepackage{amsmath}
\usepackage{amssymb}
\usepackage{booktabs}
\usepackage{color}
\usepackage{float}
\usepackage{bbding}
\usepackage{mwe}

\usepackage[pagebackref,breaklinks,colorlinks]{hyperref}

\title{Open-World Amodal Appearance Completion}

\author{
    Jiayang Ao$^{1}$, Yanbei Jiang$^{1}$, Qiuhong Ke$^{2}$, Krista A. Ehinger$^{1}$\\
    $^1$The University of Melbourne \\
    $^2$Monash University\\
}

\begin{document}

\twocolumn[{
    \maketitle
    \begin{figure}[H]
        \hsize=\textwidth 
        \centering
        \vspace*{-0.5cm}
        \includegraphics[width=\textwidth]{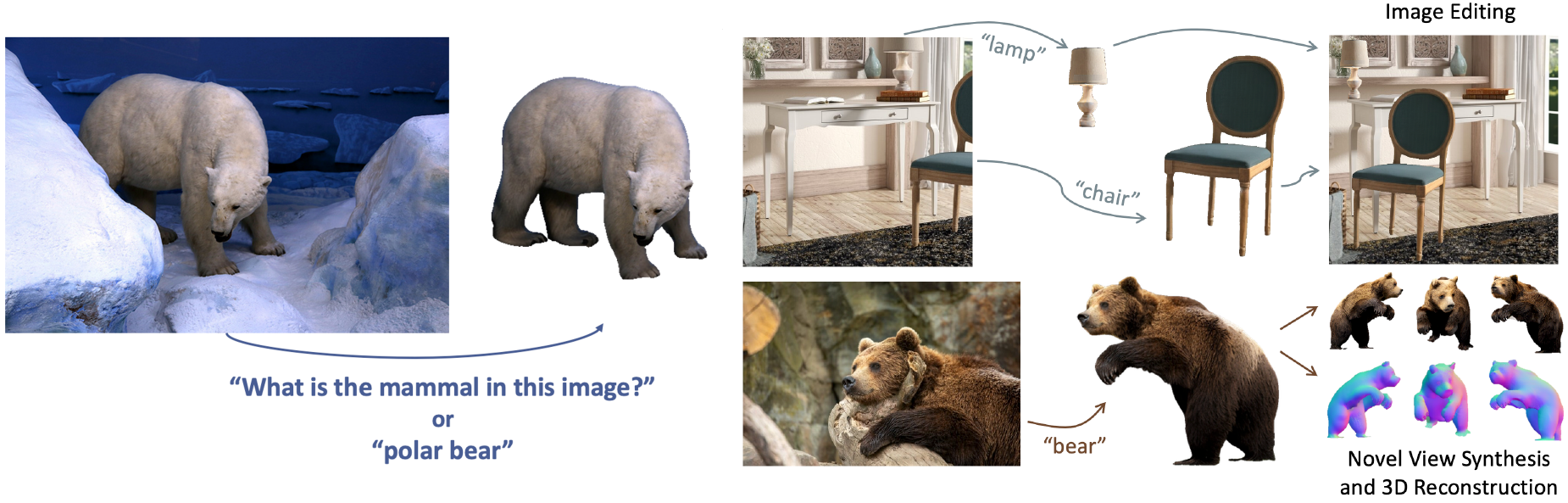}
        \caption{Examples of our open-world amodal completion using both specific (e.g., “polar bear") and abstract (e.g., “What is the mammal in this image") text queries. Our approach supports various applications, including image editing, novel view synthesis and 3D reconstruction.}
        \label{fig:first}
        \label{intrograph}
    \end{figure}}]

\maketitle
\begin{abstract}
Understanding and reconstructing occluded objects is a challenging problem, especially in open-world scenarios where categories and contexts are diverse and unpredictable. Traditional methods, however, are typically restricted to closed sets of object categories, limiting their use in complex, open-world scenes. We introduce Open-World Amodal Appearance Completion, a training-free framework that expands amodal completion capabilities by accepting flexible text queries as input. Our approach generalizes to arbitrary objects specified by both direct terms and abstract queries. We term this capability reasoning amodal completion, where the system reconstructs the full appearance of the queried object based on the provided image and language query. Our framework unifies segmentation, occlusion analysis, and inpainting to handle complex occlusions and generates completed objects as RGBA elements, enabling seamless integration into applications such as 3D reconstruction and image editing. Extensive evaluations demonstrate the effectiveness of our approach in generalizing to novel objects and occlusions, establishing a new benchmark for amodal completion in open-world settings.\footnote{The code and datasets will be released after paper acceptance.} 
\end{abstract}    
\section{Introduction}
\label{sec:intro}

Imagine taking a photograph where part of a landmark is hidden behind trees, or watching a video where an interesting object is partly occluded by other objects in the frame. What if you could automatically reconstruct the hidden parts of these objects to use them in photo editing, augmented reality (AR), or even 3D modelling? This capability, known as amodal completion~\cite{kanizsa1979organization}, allows systems to infer and generate the occluded parts of objects, providing users with full representations of partially visible items. Amodal appearance completion is critical in applications such as AR~\cite{ling2020variational,gkitsas2021panodr,pintore2022instant}, 3D reconstruction~\cite{zhan2024amodal, li20222d}, and content creation~\cite{bar2022text2live,tudosiu2024mulan,zhang2024transparent}, where intuitive, language-based interaction allows users to specify objects directly~\cite{honerkamp2024language,mo2023towards,brooks2023instructpix2pix}. Traditional amodal appearance completion methods, however, are typically constrained by fixed sets of object categories~\cite{xu2024amodal}, or necessitating extensive training and limiting their applicability in diverse and changing environments~\cite{ozguroglu2024pix2gestalt}. The real world presents a more complex challenge: completing occluded objects that can belong to any categories and exhibit diverse, unseen visual features.


To address the gap, we propose \textbf{Open-World Amodal Appearance Completion}, a framework that generalizes amodal completion to arbitrary objects specified by flexible natural language queries—a task we term \textbf{reasoning amodal completion}. Our framework supports text-based queries, allowing users to specify target object through both concrete terms and more abstract, context-dependent descriptions, as shown in~\cref{fig:first}. This flexibility introduces new possibilities for reasoning-driven amodal completion, where the model can infer and complete occluded objects based on the query alone, without predefined object categories or additional training.

Our approach is designed to be \textbf{training-free}, utilizing the extensive knowledge embedded within large, pre-trained models to deliver realistic amodal completion without requiring further data or retraining. By leveraging complementary capabilities across multiple models, our framework integrates segmentation, occlusion analysis, and progressive inpainting into a unified pipeline. Specifically, segmentation isolates the target object based on the text query, occlusion analysis identifies and resolves occluders, and inpainting reconstructs the occluded portions of the specified object with high fidelity.

The framework produces completed objects as RGBA elements, which are immediately ready for integration into downstream applications such as 3D scene reconstruction, AR, and image editing. By enabling intuitive, free-form text input and bypassing the need for retraining, we expand the application scope of amodal completion to complex, real-world scenarios where conventional closed-set methods fail.

Our primary contributions are as follows:

\begin{itemize}
\setlength{\itemsep}{6pt}

\item Open-World Framework for Amodal Completion: We propose a training-free framework capable of completing arbitrary objects specified by natural language queries, advancing amodal appearance completion beyond traditional, closed-set approaches.

\item Reasoning Amodal Completion: We introduce a reasoning-driven method that allows users to specify objects through both specific terms and abstract queries, enhancing amodal completion for intuitive human-computer interaction in real-world scenarios.

\item Adaptable Output for Enhanced Integration: Our framework outputs completed objects as RGBA elements, facilitating seamless use in AR, novel view synthesis, 3D reconstruction, and image editing.

\end{itemize}

Through extensive evaluations, including a newly collected evaluation dataset and human preference studies, we demonstrate that our framework effectively generalizes to diverse objects and complex occlusions in real-world scenes. Our approach surpasses existing methods in both quantitative and user-centered metrics, establishing a new benchmark for open-world amodal appearance completion.

\section{Related Work}
\label{sec:literature}

\textbf{Amodal Appearance Completion and Open-World Adaptability.}
Amodal completion aims to reconstruct occluded regions of objects to improve scene understanding~\cite{ao2023image}. Traditional approaches, however, are often limited to specific, predefined categories and require extensive training data. Category-specific models for vehicles~\cite{yan2019visualizing}, humans~\cite{zhou2021human,zhang2022face}, and food items~\cite{papadopoulos2019make} can effectively reconstruct occlusions within narrow classes but lack generalization to open-world settings.

Efforts to broaden amodal appearance completion include methods like Pix2gestalt~\cite{ozguroglu2024pix2gestalt}, which synthesizes occluded appearance across various objects but relies on supervised learning from large datasets, limiting adaptability. Other synthetic data-driven methods like CSDNet~\cite{zheng2021visiting} and PACO~\cite{liu2024object} have been proposed to handle complex scenes but may face challenges in adapting from synthetic to real-world environments. Self-supervised methods, such as PCNet~\cite{zhan2020self}, reduce data dependency but still require visible masks for all objects in the image, making them vulnerable to segmentation errors. PD-MC (Progressive Diffusion with Mixed Context diffusion sampling)~\cite{xu2024amodal} partially leverages pre-trained models for broader generalization but remains constrained by its reliance on predefined categories. 

Therefore, we need a robust framework that can handle arbitrary objects in the open world without relying on specific categories or extensive data annotations.

\textbf{Reasoning-Driven and Adaptable Amodal Completion.} Recent vision-language models (VLM) allow users to specify objects contextually, enriching human-computer interaction in vision tasks. General-purpose models like CogVLM~\cite{wang2023cogvlm} and VisionLLM~\cite{wang2024visionllm} perform well in object recognition and language-guided tasks but struggle with reasoning about occlusions. LISA~\cite{lai2024lisa} offers improved reasoning for identifying visible objects through language input, yet it falls short when dealing with occluded regions. However, amodal completion requires reasoning beyond the visible to reconstruct these occlusions, providing a full representation of objects that benefits various applications. Increasingly, RGBA output formats are seen as vital for downstream applications, as they support flexible blending and realistic compositing~\cite{zhang2024transparent}. Datasets like MULAN~\cite{tudosiu2024mulan} underscore the importance of RGBA instances in amodal tasks. Altogether, there is a clear need for reasoning-driven amodal completion that adapts dynamically across open-world objects while producing adaptable RGBA outputs.

To address the limitations of existing approaches, our framework introduces a training-free method for open-world amodal appearance completion, accommodating flexible, language-guided queries to identify and complete arbitrary objects. By integrating RGBA outputs, our method ensures adaptable representations that enhance its applicability across diverse downstream tasks.

\begin{figure*}[ht]
  \centering
   \includegraphics[width=0.9\linewidth]{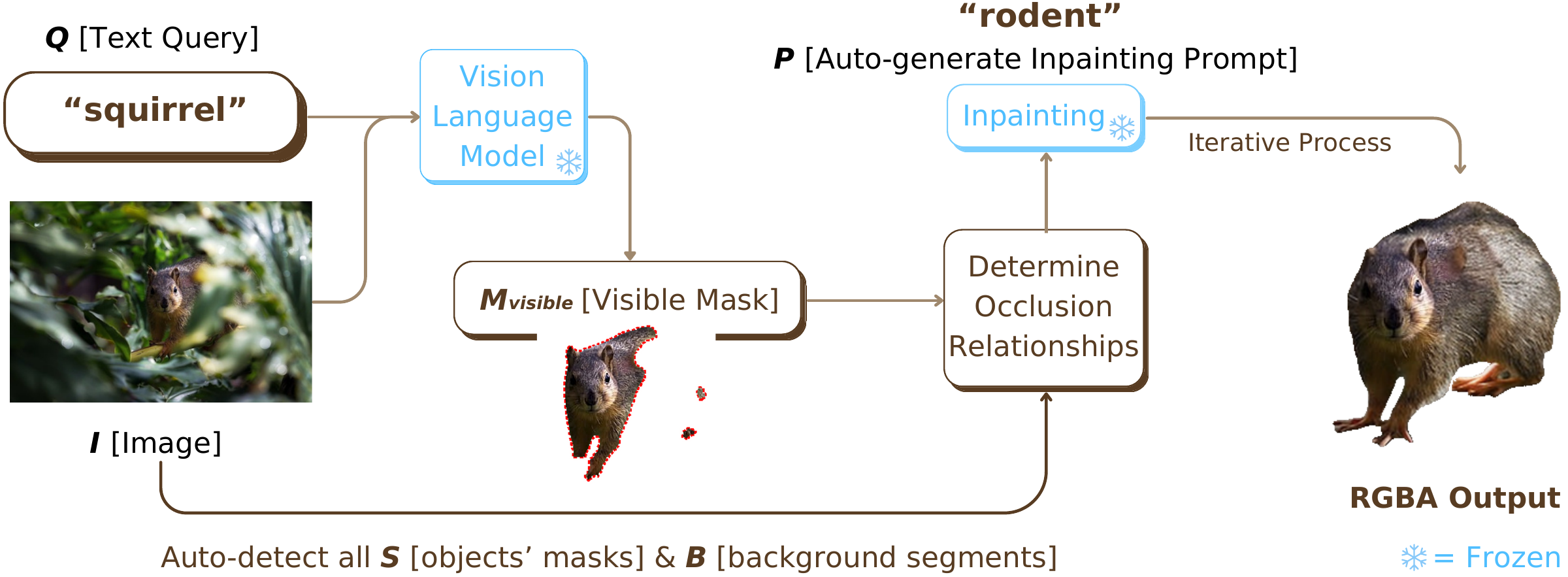}
   \caption{Overview of our framework. Starting with a text query, a VLM generates a visible mask to locate the target object in the input image. The framework then identifies all objects and background segments for occlusion analysis. An auto-generated prompt guides the inpainting model, which iteratively reconstructs the occluded object to produce a transparent RGBA amodal completion output.}
   \label{fig:method}
   \vspace*{-0.3cm}
\end{figure*}

\section{Methodology}
\label{sec:method}

We introduce a novel reasoning-driven approach for reconstructing occluded objects across diverse, unrestricted categories. By leveraging the flexibility of natural language, users can specify target object through both concrete terms and abstract descriptions, allowing amodal completion that generalizes beyond the constraints of closed-set methods. This adaptability is achieved without additional training, allowing it to effectively handle open-world scenes and unknown objects. \cref{fig:method} presents an overview of our end-to-end inference pipeline, illustrating how flexible text-based input is processed to yield high-quality, transparent RGBA outputs. Our approach isolates the completed object as a layerable element, enabling straightforward integration in compositional applications.

In Section~\ref{sec:textquery_segmentation}, we describe how the natural language query is processed to generate precise visible segmentation object mask. Section~\ref{sec:occlusion_analysis} details our occlusion analysis strategy, which entails identifying and masking occluders that occlude the target object. In Section~\ref{sec:prompt_selection}, we present a heuristic method for selecting the inpainting text prompt, enabling targeted reconstruction of occluded regions based on the visible part of object. Section~\ref{sec:iterative_inpainting} outlines the iterative inpainting approach that progressively completes the occluded object and produces an RGBA output (an RGB image with an alpha mask).

\textbf{Definitions.} Let \( I \in \mathbb{R}^{H \times W \times 3} \) represent the input image of height \( H \) and width \( W \), with \( Q \) as a natural language query that specifies the target object in the scene. The query \( Q \) may explicitly name the object (e.g., “polar bear”) or describe it in context (e.g., “the animal in this image”). The output of the framework is an amodal completion \( C \in \mathbb{R}^{H \times W \times 4} \), an RGBA representation where the first three channels represent RGB color values, and the fourth channel (alpha) encodes transparency for adaptable layering in downstream applications. The amodal completion process includes generating an initial visible mask \( M_{\text{visible}} \in \{0, 1\}^{H \times W} \), identifying the occluder mask \( M_{\text{occ}} \), and iteratively updating these masks to achieve a fully reconstructed representation of the target object with background transparency.

\subsection{Text Query Interpretation and Image Segmentation}
\label{sec:textquery_segmentation}

One primary challenge in amodal completion is accurately isolating the target object in diverse, complex scenes, especially when the object is described in natural language. However, robust amodal completion also requires a comprehensive segmentation of all visible elements within the image, as any other object could potentially act as an occluder.

Given an input image \( I \in \mathbb{R}^{H \times W \times 3} \) and a natural language query \( Q \) describing the target object, we first generate an initial mask \( M_{\text{visible}} \in \{0, 1\}^{H \times W} \) that identifies the visible regions of \( I \) corresponding to the target object. This mask is derived from a pre-trained language-grounded vision model~\cite{lai2024lisa} that aligns \( Q \) with spatial areas relevant to the query, allowing the system to handle both specific and abstract descriptions. This initial mask \( M_{\text{visible}} \) provides a foundation for understanding occlusion relationships with other elements in the scene.

\textbf{Context-Aware Object Segmentation.} While \( M_{\text{visible}} \) offers an initial visible boundary for the target object, it is crucial to identify other objects and background areas within the image, as these may act as occluders. We employ the automatic image annotation system proposed in~\cite{ren2024grounded} to automatically identify all nameable objects in the image. The system first extracts a set of contextual tags \( T = \{t_1, t_2, \dots, t_n\} \) from the image with a pre-trained open-set image tagging model~\cite{huang2023open}, where each tag \( t_i \) denotes a class label for a recognizable visual feature in \( I \) (e.g., “cat,” “plate,” or “bus”). Using \( T \) and \( I \) as inputs, the system leverage a pre-trained open-set object detector~\cite{liu2024grounding} combine with the Segment Anything model~\cite{kirillov2023segment} to generate a set of segmentation masks \( S = \{S_1, S_2, \dots, S_m\} \), where each mask \( S_i \) represents a visible objects within \( I \). This context-aware segmentation framework enables the pipeline to delineate each visible object, including the target, with well-defined boundaries, ensuring all potential occluders are identified and isolated.

\textbf{Handling Background Regions and Unknown Objects.} In addition to identifiable objects, many scenes contain visually ambiguous or hard-to-describe areas, such as background elements, blurry or unidentifiable objects, clutter, and textures. Traditional segmentation methods may overlook these regions because they aren't associated with object category labels, but these unlabelled (or “background”-labelled) regions can be occluders of a target object. Our approach handles these areas through an segmentation process which divides these “missing” areas into distinct segments without relying solely on object-category-driven segmentation.

After generating \( S \), we identify unsegmented areas within \( I \), defined as \( B \subset I \), which consists of ambiguous background elements or hard-to-describe areas that may still interact with the target object through occlusion. These unsegmented regions pose a problem for occlusion analysis because they may be large and contain multiple objects or surfaces, only some of which may occlude the target object. To process these areas, we apply a sequence of morphological operations that refine and partition \( B \) into a collection of background segments. Specifically, we first generate a binary mask from \( B \), where unsegmented pixels are assigned a value of 1 and all other pixels a value of 0. This binary mask is then eroded using a structured element to sharpen boundaries and separate loosely connected areas, preventing overlap with object segments in \( S \). Following erosion, we apply dilation to the refined areas, which helps to re-expand and consolidate adjacent regions into distinct connected components, resulting in a set of background segments \( B = \{B_1, B_2, \dots, B_k\} \), where:
\begin{equation}
B_j = \text{Morph}(I - \bigcup_{i=1}^m S_i), \quad \forall j \in \{1, 2, \dots, k\}
\end{equation}
Here, \( \text{Morph} \) denotes the combined effect of the morphological operations applied to partition ambiguous regions into isolated segments. 

This process allows our method of consider a wider range of occlusion conditions, such as target objects occluded by background (e.g., buildings partially hidden by undergrowth) and cases where the occluding objects are blurry, ambiguous, or difficult to recognize.

\subsection{Occlusion Analysis}
\label{sec:occlusion_analysis}

To reconstruct the missing parts of the target object, it is essential to identify where the object is occluded -- these occluded areas are where completion is required. Our method generates an occluder mask for the target object that highlights the occluded areas. This mask is used to guide the inpainting process.

Given an input image \( I \in \mathbb{R}^{H \times W \times 3} \), the set of object masks \( S = \{S_1, S_2, \dots, S_m\} \) and the set of background segments \( B = \{B_1, B_2, \dots, B_k\} \), we define an occluder mask \( M_{\text{occ}} \in \{0, 1\}^{H \times W} \) that aggregates all segments which occlude the target object.

\textbf{Occluder Identification through Spatial Analysis.} 
The occluder mask \( M_{\text{occ}} \) is derived by first evaluating the spatial relationships between the target’s visible mask \( M_{\text{visible}} \) and the set of segmented regions in \( S \cup B \), where each segment potentially overlaps or interacts with the target. Our framework employs InstaOrderNet~\cite{lee2022instance}, a pre-trained model for occlusion orderings that takes pairwise segmentation masks and an image patch as input to determine the occlusion order between segment pairs, without requiring class labels. This approach enables identification of which segments within \( S \cup B \) occlude the target object.

For each segment \( S_i \) or \( B_j \) in the scene, InstaOrderNet computes a binary occlusion indicator, assigning a value of \( \text{occ} = 1 \) if the segment occludes the target and \(0\) otherwise. The occluder mask \( M_{\text{occ}} \) is thus initialized as the union of all occluding segments in \( S \) and \( B \) identified by InstaOrderNet, formally expressed as:
\begin{equation}
M_{\text{occ}} = \bigcup_{\substack{i = 1 \\ \text{occ}_{i} = 1}}^{m} S_i \cup \bigcup_{\substack{j = 1 \\ \text{occ}_{j} = 1}}^{k} B_j
\end{equation}
where \( \cup \) denotes pixel-wise union.

\textbf{Boundary-Aware Occlusion.}
Scenes often present cases where the target object touches the image boundary, in which case amodal completion requires image expansion~\cite{xu2024amodal}. Thus we apply an adaptive refinement strategy that expands \( M_{\text{occ}} \) along the target’s boundary regions, guided by iterative boundary adjustments.

The process begins by checking for boundary contacts in \( M_{\text{visible}} \), where the mask contacts any edge of \( I \). For boundary-touching regions, we apply a dilation operation to the occluder mask, increasing its coverage along the contacted edges. Let \( E \subset \{ \text{top, bottom, left, right} \} \) denote the set of edges contacted by \( M_{\text{visible}} \), and let \( d \) represent a structured dilation operation. We update \( M_{\text{occ}} \) as follows:
\begin{equation}
M_{\text{occ}} \leftarrow M_{\text{occ}} \cup \left( d(M_{\text{visible}}) \cap \bigcup_{e \in E} \text{edge}_{e} \right)
\end{equation}
where \( \text{edge}_{e} \) refers to the pixels along the boundary edge \( e \) of the image, and \( \cap \) denotes pixel-wise intersection. This expansion is applied iteratively, refining the occluder mask’s coverage until the mask stabilizes or the set of edges \( E \) is fully dilated to ensure coverage for the amodal completion process.

\subsection{Prompt and Image Refinement for Inpainting}
\label{sec:prompt_selection}

Because our system is designed to handle a wide range of natural language queries (e.g., “the animal in this image”) we cannot assume that the query alone will be sufficient to guide the inpainting step. We therefore combine the query with the image context to refine the inpainting prompt.

Given an input image \( I \in \mathbb{R}^{H \times W \times 3} \), visible mask \( M_{\text{visible}} \), and a natural language query \( Q \), we aim to produce an optimal inpainting prompt \( P \) that effectively guides the reconstruction process. Using a CLIP-based similarity comparison~\cite{radford2021learning}, we match the visible target object in \( I_{\text{target}} = I \odot M_{\text{visible}} \)  (where \(\odot\) denotes element-wise multiplication) with candidate descriptors from the image tags \( T \cup Q \), selecting the descriptor that best matches the visible attributes of the target object. This similarity \( S(t_i) \) for each candidate \( t_i \in T \cup Q \) is computed as:
\begin{equation}
P = \arg\max_{t_i \in T \cup \{Q\}} \, \text{CLIP}(I_{\text{target}}, t_i)
\end{equation}
where CLIP extracts features for visual-text alignment.

To further isolate the target, regions outside \( M_{\text{visible}} \) in \( I_{\text{target}} \) are swapped with a clean background \(I_{\text{bkgd}}\), inspired by the background isolation approach in~\cite{xu2024amodal}. This prevents irrelevant areas from influencing the prompt. Our background swapping step ensures the prompt \( P \) focuses solely on the target object, improving inpainting precision. The prompt \( P \) is then used as a conditioning input for the inpainting model.


\subsection{Iterative Inpainting and Amodal Completion}
\label{sec:iterative_inpainting}

To achieve realistic amodal completion, our framework performs iterative masked inpainting to reconstruct occluded parts of the target object, terminating based on the occluder mask stability or a set iteration limit. After inpainting completes, a final seamless blending step merges the original visible regions with the reconstructed occluded parts, ensuring alignment with the original image and producing a transparent RGBA output that excludes unrelated background or extraneous elements.

Given an input image \( I \in \mathbb{R}^{H \times W \times 3} \), visible mask \( M_{\text{visible}} \), occluder mask \( M_{\text{occ}} \), and inpainting prompt \( P \), the reconstruction process begins by isolating the target object from the surrounding scene. We initialize an inpainting target \( I_{\text{target}} \) as follows:
\begin{equation}
I_{\text{target}} = I \odot M_{\text{visible}} + (1 - M_{\text{visible}}) \odot I_{\text{bkgd}}
\end{equation}
where \( I_{\text{bkgd}} \) represents a clean background (similar to the natural, context-free background in~\cite{xu2024amodal}), ensuring that only the target object remains visible for inpainting.

The iterative process begins by setting \( I_{\text{inpaint}}^{(0)} = I_{\text{target}} \), where \( I_{\text{inpaint}}^{(t)} \) denotes the inpainting result at iteration \( t \). At each step \( t \), a pre-trained inpainting model~\cite{rombach2022high} \( \phi \) fills occluded regions in \( I_{\text{inpaint}}^{(t)} \) based on \( P \) and the updated occluder mask \( M_{\text{occ}}^{(t)} \):
\begin{equation}
I_{\text{inpaint}}^{(t+1)} = \phi(I_{\text{inpaint}}^{(t)}, M_{\text{occ}}^{(t)}, P)
\end{equation}

In each iteration, the occluder mask \( M_{\text{occ}}^{(t+1)} \) is refined to include newly inpainted regions, while an updated amodal segmentation mask \( M_{\text{amodal}} \) is computed to accurately bound the target object. This iterative refinement continues until reaching the final inpainted output \( I_{\text{inpaint}}^{(T)} \), following the approach in~\cite{xu2024amodal}.

\textbf{Adaptive Termination for Efficiency.} To ensure computational efficiency, we apply an adaptive termination criterion based on occluder mask stability. If the pixel-wise difference between consecutive occluder masks,
\begin{equation}
\Delta M_{\text{occ}}^{(t)} = \| M_{\text{occ}}^{(t+1)} - M_{\text{occ}}^{(t)} \|_{L_1}
\end{equation}
falls below a threshold \( \epsilon \) or the inpainting process reaches a maximum number of iterations \( T \), the iteration stops, yielding the final inpainted result \( I_{\text{inpaint}}^{(T)} \).

\textbf{Blending for Seamless Integration.} Once inpainting is complete, a final alpha blending step is applied to achieve a smooth transition between newly reconstructed and original visible region of the object. This blending operates from the interior toward a "transition region" of a predefined width along the boundary of the visible region. Specifically, the center of the original visible region maintains an alpha value of 1, and areas outside the visible region are set to 0. Within the transition region, alpha values gradually decrease from 1 to 0. The blended image \( I_{\text{blend}} \) is defined as:
\begin{equation}
I_{\text{blend}} = M_{\alpha} \cdot I_{\text{inpaint}}^{(T)} + (1 - M_{\alpha}) \cdot I_{\text{orig}}.
\vspace*{-0.3cm}
\end{equation}
\begin{figure*}[ht]
  \centering
   \includegraphics[width=\linewidth]{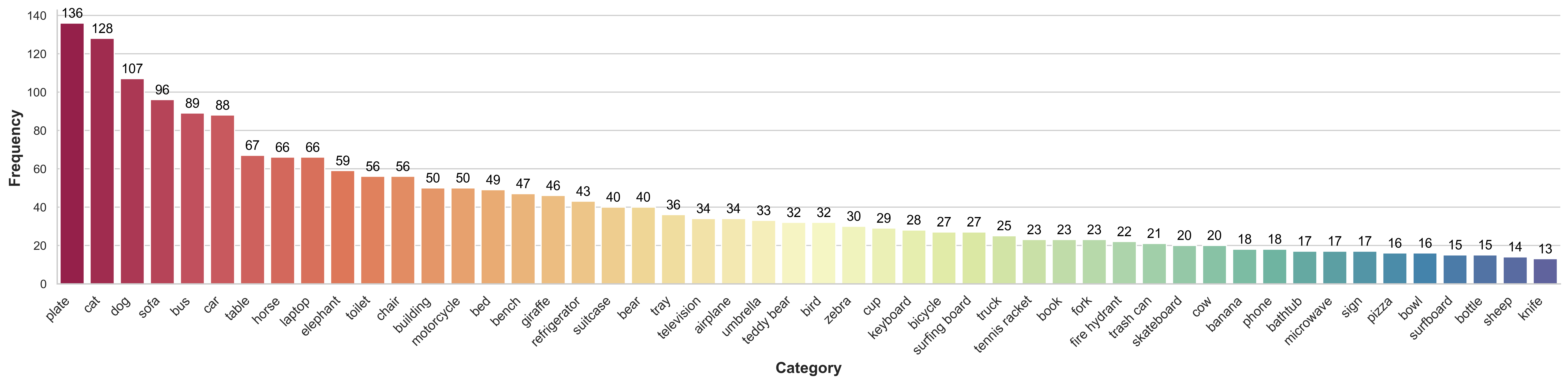}
   \caption{Distribution of the top 50 most frequent categories in the our evaluation dataset.}
   \label{fig:distribution}
   \vspace*{-0.4cm}
\end{figure*}

\textbf{RGBA Output for Versatile Use.} The final output is an RGBA image \( C \in \mathbb{R}^{H \times W \times 4} \), with the first three channels corresponding to the RGB values of \( I_{\text{blend}} \) and the fourth channel \( M_{\text{amodal}} \) representing the binary amodal segmentation mask.


\section{Experiments}
\label{sec:experiment}

\subsection{Datasets Collection}

\begin{table}[ht]
\setlength{\tabcolsep}{3.5pt}
\centering
\footnotesize
\begin{tabular}{lcccc|c}
\toprule
\multicolumn{1}{c}{Source}                                     & \multicolumn{1}{c}{VG} & \multicolumn{1}{c}{COCO-A} & \multicolumn{1}{c}{Free Image} & \multicolumn{1}{c|}{Laion} & \multicolumn{1}{c}{Total} \\ \hline
\# Img                                                          & 1234                             & 751                       & 228                            & 166                        & 2379                      \\
\# Instances                                                    & 1294                             & 845                       & 249                            & 177                        & 2565                      \\
\# Classes                                                   & 269                              & 256                       & 147                            & 97                         & 553                       \\
\# Img w/ 1 occludee & 1184                             & 663                       & 211                            & 156                        & 2214                      \\
\# Img w/ 1+ occludees                                          & 50                               & 88                        & 17                             & 10                         & 165                       \\ \bottomrule
\end{tabular}
\caption{Composition of the evaluation dataset. The broad variety of classes across different sources presents a challenge for open-world amodal appearance completion.}
\label{tab:datacollect}
\vspace{-0.5cm}
\end{table}

Existing datasets for amodal appearance completion often have limitations, either due to constraints on a limited set of object categories or reliance on synthetic occlusion data~\cite{zhu2017semantic, ehsani2018segan,follmann2019learning,hu2019sail,li2023muva,tudosiu2024mulan}, which fall short of capturing the complexity of real-world scenarios. To provide a comprehensive evaluation for the open-world amodal appearance completion challenge, we gathered a diverse dataset with natural occlusions and a broad range of object categories across varied scenes.

Our evaluation dataset draws from four sources: COCO-A~\cite{zhu2017semantic}, Visual Genome (VG)~\cite{krishna2017visual}, LAION~\cite{schuhmann2021laion}, and copyright-free images from publicly accessible websites. Each source brings unique strengths to our evaluation dataset. COCO-A and VG provide natural scenes containing occluded objects real-world environments. Given that COCO-A was initially designed for semantic segmentation, not all images feature object-specific occlusions. Thus, we applied a filtering process to COCO-A, retaining only those images in which objects are  occluded by other objects.\footnote{Since the COCO-A dataset was originally developed for semantic segmentation, it focuses on semantic-level occlusions—such as occluded background elements—rather than solely object-level occlusions. Consequently, not all images contain relevant object occlusions; we removed these from our COCO-A selection. Details of the selection process are provided in the appendix.}. The addition of LAION and free-image sources introduces the diversity of internet-sourced images, covering a wide range of settings, lighting, and complex occlusion types characteristic of unconstrained environments.

The resulting dataset consists of 2379 images spanning 553 distinct target object classes (see~\cref{tab:datacollect}). Three human annotators collected images containing occluded objects and provided a class label for each occluded object. \cref{fig:distribution} presents the top 50 most frequent occluded object categories, illustrating the distribution across classes. This diversity ensures that methods can be evaluated across a wide range of object categories and occlusion scenarios, resulting in a challenging testbed for open-world amodal appearance completion.

\subsection{Implementation and Evaluation Metrics}
\textbf{Implementation Details.} Our framework leverages publicly available models without additional training or fine-tuning to deliver high-quality amodal completions. The main models utilized include the Stable Diffusion v2 inpainting model~\cite{rombach2022high} and the LISA-13B-llama2-v1 model~\cite{lai2024lisa}, both publicly available.

All experiments were conducted on an NVIDIA A100 GPU. For comparison methods, we ran inference using their publicly available pre-trained models with default settings. Our framework’s design maintains computational efficiency by limiting the inpainting process to a maximum of three iterations per object; further details on our configurations are provided in the appendix.

\textbf{Evaluation metrics.} Evaluating amodal completion on natural images with real-world occlusions presents unique challenges, as the ground-truth appearance of occluded regions is inherently unavailable. Thus, we use a combination of human evaluation and quantitative metrics. We center our evaluation on human assessment, which is critical for judging the subjective quality of amodal completion outputs. 

We use CLIP score~\cite{radford2021learning} to measure how well the amodal completions align with the class labels. Specifically, we compare the amodal completion of each object with the ground-truth class label associated with the object.

\begin{figure*}[ht]
  \centering
   \includegraphics[width=0.87\linewidth]{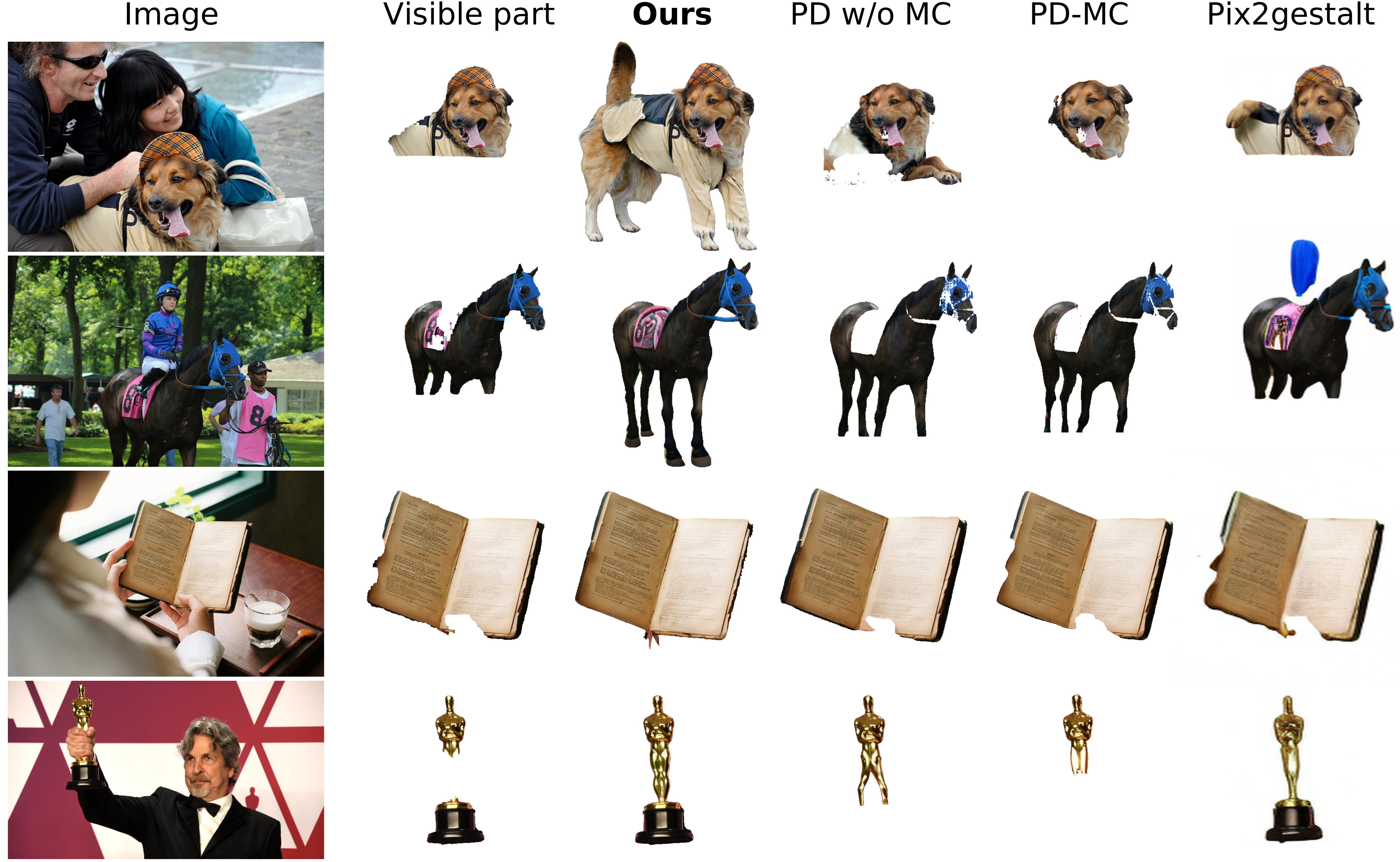}
   \caption{Visual comparisons of amodal completions across different methods: Ours consistently outperforms others in terms of realism, handling complex occlusions, and producing plausible completions. Top to bottom: examples from VG, COCO-A, free image, LAION.}
   \label{fig:prediction}
   \vspace*{-0.4cm}
\end{figure*}

We also include appearance quantitative metrics. As only the visible parts of each object are available, these metrics offer a rough reference of completion quality. We compute LPIPS~\cite{zhang2018unreasonable} for visual consistency, VGG16 feature similarity~\cite{gatys2016image} for semantic consistency, and SSIM~\cite{wang2004image} for structural consistency—all by comparing the visible part of the object with the completed version.

\subsection{Comparisons with Other Methods} 
Since amodal completion is inherently subjective, human evaluation serves as a reliable measure of realism and completion quality. We conducted a preference study on Prolific crowdsourcing platform\footnote{\url{https://www.prolific.com/}} with 180 participants, comparing our method to Pix2gestalt~\cite{ozguroglu2024pix2gestalt}, PD-MC~\cite{xu2024amodal}, and a baseline PD w/o MC (PD-MC without Mixed Context Diffusion). Participants were presented with the original image and four completed versions, each randomized to avoid positional bias. They were instructed to choose the completion that appeared most realistic, considering both visible and occluded regions. To ensure data reliability, “gold standard” questions with straightforward choices were included, and only participants who passed 75\% of these checks were retained in the final analysis.

\begin{table}[t]
\centering
\footnotesize
\setlength{\tabcolsep}{5pt}
\begin{tabular}{cccccc}
\toprule
\begin{tabular}[c]{@{}c@{}}Dataset/\\ Method\end{tabular} & VG               & COCO-A            & \begin{tabular}[c]{@{}c@{}}Free\\ Image\end{tabular} & LAION            & Overall          \\ \hline
PD w/o MC                                                     & 16.62\%                                                         & 15.38\%                                                         & 15.80\%                                                                       & 11.49\%                                                         & 15.78\%                                                         \\
PD-MC                                                      & 15.51\%                                                         & 14.20\%                                                         & 12.99\%                                                                       & 9.42\%                                                          & 14.41\%                                                         \\
Pix2gestalt                                               & 26.56\%                                                         & 31.24\%                                                         & 21.29\%                                                                       & 31.83\%                                                         & 27.95\%                                                         \\
Ours                                                      & \cellcolor[HTML]{E6F2E6}{\color[HTML]{000000} \textbf{41.32\%}} & \cellcolor[HTML]{E6F2E6}{\color[HTML]{000000} \textbf{39.17\%}} & \cellcolor[HTML]{E6F2E6}{\color[HTML]{000000} \textbf{49.93\%}}               & \cellcolor[HTML]{E6F2E6}{\color[HTML]{000000} \textbf{47.27\%}} & \cellcolor[HTML]{E6F2E6}{\color[HTML]{000000} \textbf{41.86\%}} \\ \bottomrule
\end{tabular}
\caption{Human preference study across four different sources. The table shows the percentage of participants who preferred each method's amodal completion results.}
\label{tab:preference}
\vspace{-0.3cm}
\end{table}

\begin{figure}[t]
  \centering
   \includegraphics[width=\linewidth]{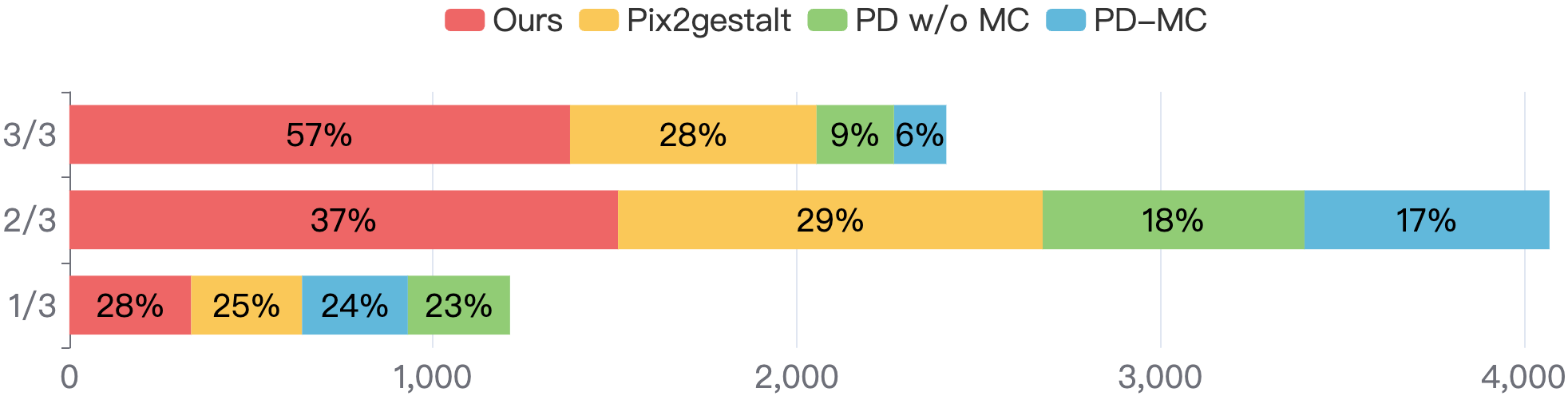}
   \caption{Model preference of human evaluators by agreement levels. X-axis shows the number of images. 3/3 denotes full agreement among three evaluators per image, 1/3 indicates no consensus. “Ours" shows the strongest consensus on completion quality.}
   \label{fig:agreement}
   \vspace{-0.3cm}
\end{figure}

Our method achieved the highest overall preference in the evaluation dataset, as shown in~\cref{tab:preference}. Across all four sources, our approach consistently ranked first, demonstrating robustness in handling diverse occlusions across open-world scenarios. The agreement levels among participants further highlight our performance. As illustrated in~\cref{fig:agreement}, in cases where all three participants agreed on a single preferred completion (3/3), our method was selected 57\% of the time, followed by Pix2gestalt at 28\%, showing the perceptual quality and realism of our results. Even in cases of partial consensus (2/3) or no majority (1/3), our approach still retained a lead, indicating a consistent preference even when there was divergence in evaluator opinion.

\cref{fig:prediction} shows visual comparisons of our method with others on challenging mutual occlusions across open-world scenarios. In the dog and horse images, which involve extensive outpainting, only our approach reconstructs the full body accurately. Similarly, in the book and Oscar statuette examples, ours achieves coherent and realistic completions, preserving structural details that other methods miss. This highlights our method’s strength in handling complex objects and challenging occlusions with superior realism.

\textbf{Quantitative Metrics.} We assessed completion quality using quantitative metrics: CLIP score for class relevance, and LPIPS, VGG16 feature similarity, and SSIM for appearance quality. As shown in the~\cref{tab:quantitative}, our method achieved the highest CLIP score, aligning well with ground-truth object categories and showcasing strong relevance. In appearance metrics, our method also outperformed others, indicating superior visual, semantic and structural consistency. Failure cases where no amodal completion is generated result in LPIPS of 1 and feature similarity/SSIM of 0. While these appearance quantitative metrics provide insight, they are provided for reference only, as the amodal appearance ground-truth is not available.


\begin{table}[]
\centering
\footnotesize
\setlength{\tabcolsep}{1.5pt} 
\begin{tabular}{ccccccc}
\toprule
\multicolumn{1}{l}{} & & \multicolumn{1}{c}{\cellcolor[HTML]{EFEFEF}\begin{tabular}[c]{@{}c@{}}Class \\ Relevance\end{tabular}} & \multicolumn{1}{c}{\cellcolor[HTML]{EFEFEF}\begin{tabular}[c]{@{}c@{}}Visual \\ Consistency\end{tabular}} & \multicolumn{1}{c}{\cellcolor[HTML]{EFEFEF}\begin{tabular}[c]{@{}c@{}}Semantic \\ Consistency\end{tabular}} & \multicolumn{1}{c}{\cellcolor[HTML]{EFEFEF}\begin{tabular}[c]{@{}c@{}}Structural \\ Consistency\end{tabular}} \\
Dataset & \multicolumn{1}{c}{Method} & \multicolumn{1}{c}{↑ CLIP} & \multicolumn{1}{c}{↓ LPIPS} & \multicolumn{1}{c}{\begin{tabular}[c]{@{}c@{}}↑ Feature \\ Similarity\end{tabular}} & \multicolumn{1}{c}{↑ SSIM} \\ \hline
\multirow{4}{*}{\textbf{VG}}                                                     & PD w/o MC                   & 28.254                                                                          & 0.586                                                                             & 0.406                                                                               & 0.459                                                                                 \\
& PD-MC                    & 28.367& 0.578& 0.413& 0.463\\
& Pix2gestalt             & 27.672& 0.429& 0.554& 0.726\\
& Ours                    & \cellcolor[HTML]{E6F2E6}{\color[HTML]{000000} \textbf{28.470}}& \cellcolor[HTML]{E6F2E6}{\color[HTML]{000000} \textbf{0.310}}& \cellcolor[HTML]{E6F2E6}{\color[HTML]{000000} \textbf{0.658}}& \cellcolor[HTML]{E6F2E6}{\color[HTML]{000000} \textbf{0.732}}\\ \hline
\multirow{4}{*}{\textbf{COCO-A}}                                                  & PD w/o MC                   & 27.426                                                                          & 0.673                                                                             & 0.318                                                                               & 0.381                                                                                 \\
& PD-MC                    & 27.383& 0.664                                                                             & 0.328& 0.382\\
& Pix2gestalt             & 26.998& 0.471& 0.524& 0.695\\
& Ours& \cellcolor[HTML]{E6F2E6}{\color[HTML]{000000}\textbf{27.612}}& \cellcolor[HTML]{E6F2E6}{\color[HTML]{000000}\textbf{0.351}}& \cellcolor[HTML]{E6F2E6}{\color[HTML]{000000}\textbf{0.609}}& \cellcolor[HTML]{E6F2E6}{\color[HTML]{000000}\textbf{0.718}}\\ \hline
\multirow{4}{*}{\textbf{\begin{tabular}[c]{@{}c@{}}Free \\ Images\end{tabular}}} & PD w/o MC                   & 28.190                                                                          & 0.730                                                                             & 0.268                                                                               & 0.305                                                                                 \\
& PD-MC& 28.333& 0.720& 0.279& 0.309\\
& Pix2gestalt             & 27.621& 0.393& 0.613& 0.732\\
& Ours                    & \cellcolor[HTML]{E6F2E6}{\color[HTML]{000000}\textbf{28.652}}& \cellcolor[HTML]{E6F2E6}{\color[HTML]{000000}\textbf{0.269}}& \cellcolor[HTML]{E6F2E6}{\color[HTML]{000000}\textbf{0.698}}& \cellcolor[HTML]{E6F2E6}{\color[HTML]{000000}\textbf{0.753}}\\ \hline
\multirow{4}{*}{\textbf{LAION}}                                                  & PD w/o MC                   & 27.479                                                                          & 0.695                                                                             & 0.295                                                                               & 0.354                                                                                 \\
& PD-MC& 27.573& 0.692& 0.299& 0.346\\
& Pix2gestalt             & 27.260& 0.467& 0.527& 0.691\\
& Ours& \cellcolor[HTML]{E6F2E6}{\color[HTML]{000000}\textbf{28.123}}& \cellcolor[HTML]{E6F2E6}{\color[HTML]{000000}\textbf{0.319}}& \cellcolor[HTML]{E6F2E6}{\color[HTML]{000000}\textbf{0.657}}& \cellcolor[HTML]{E6F2E6}{\color[HTML]{000000}\textbf{0.751}}\\ \hline
\multirow{4}{*}{\textbf{Overall}} & PD w/o MC & 27.922 & 0.636 & 0.356 & 0.411 \\
& PD-MC & 27.984 & 0.628 & 0.364 & 0.413 \\
& Pix2gestalt & 27.417 & 0.442 & 0.548 & 0.714 \\
& Ours & \cellcolor[HTML]{E6F2E6}{\color[HTML]{000000}\textbf{28.181}} & \cellcolor[HTML]{E6F2E6}{\color[HTML]{000000}\textbf{0.320}} & \cellcolor[HTML]{E6F2E6}{\color[HTML]{000000}\textbf{0.646}} & \cellcolor[HTML]{E6F2E6}{\color[HTML]{000000}\textbf{0.731}} \\ \bottomrule
\end{tabular}
\caption{Quantitative comparisons across datasets. Our method consistently outperforms other approaches, achieving the highest scores in CLIP, LPIPS, and Feature Similarity, while maintaining the lowest LPIPS, indicating superior visual and perceptual fidelity in open-world amodal appearance completion.}
\label{tab:quantitative}
\vspace*{-0.4cm}
\end{table}

\subsection{Ablation Studies} 
\label{subsec:ablation}

\textbf{Inpainting Prompt Variants.} We tested three prompt configurations: (1) \( Q \) (text query only), (2) \( T \) (auto-generated tags only), and (3) \( T \cup Q \) (combined tags and query). As shown in~\cref{tab:ablation}, the improvement in all appearance quantitative metrics for \( T \cup Q \) indicates its effectiveness in visual, semantic and structural consistency. In our evalution dataset, the \( Q \) prompt achieves the highest CLIP score, because it directly aligns with ground-truth target object class labels. However, in practical scenarios, \( Q \) may consist of abstract or context-dependent queries rather than specific class labels, making the combined \( T \cup Q \) approach more effective by incorporating both object-specific and contextual details.

\begin{table}[ht]
\centering
\footnotesize
\setlength{\tabcolsep}{1.2pt} 
\begin{tabular}{cc|cccc}
\toprule
\multicolumn{2}{c|}{Ablation} & \cellcolor[HTML]{EFEFEF}\begin{tabular}[c]{@{}c@{}}Class \\ Relevance\end{tabular} & \cellcolor[HTML]{EFEFEF}\begin{tabular}[c]{@{}c@{}}Visual \\ Consistency\end{tabular} & \cellcolor[HTML]{EFEFEF}\begin{tabular}[c]{@{}c@{}}Semantic \\ Consistency\end{tabular} & \cellcolor[HTML]{EFEFEF}\begin{tabular}[c]{@{}c@{}}Structural \\ Consistency\end{tabular} \\
\begin{tabular}[c]{@{}c@{}}Inpainting \\ Prompt\end{tabular} & \begin{tabular}[c]{@{}c@{}}Background\\ Segments\end{tabular} & ↑ CLIP & ↓ LPIPS & \begin{tabular}[c]{@{}c@{}}↑ Feature \\ Similarity\end{tabular} & ↑ SSIM \\ \hline
\( Q \) & \Checkmark &  \cellcolor[HTML]{E6F2E6}{\color[HTML]{000000}\textbf{28.563}} & 0.327 & 0.633 & 0.724 \\
\( T \) & \Checkmark & 28.043 & 0.324 & 0.636 & 0.725 \\
\( T \cup Q \) & \XSolid & 28.071 & 0.333 & 0.620 & 0.713 \\
\( T \cup Q \) & \Checkmark & 28.181 &  \cellcolor[HTML]{E6F2E6}{\color[HTML]{000000}\textbf{0.320}} &  \cellcolor[HTML]{E6F2E6}{\color[HTML]{000000}\textbf{0.646}} &  \cellcolor[HTML]{E6F2E6}{\color[HTML]{000000}\textbf{0.731}} \\ \bottomrule
\end{tabular}
\caption{Ablation study results on the full evaluation dataset for prompt variants and background segmentation effects across class relevance and appearance quantitative metrics.}
\label{tab:ablation}
\vspace*{-0.2cm}
\end{table}

\begin{figure}[ht]
  \centering
   \includegraphics[width=\linewidth]{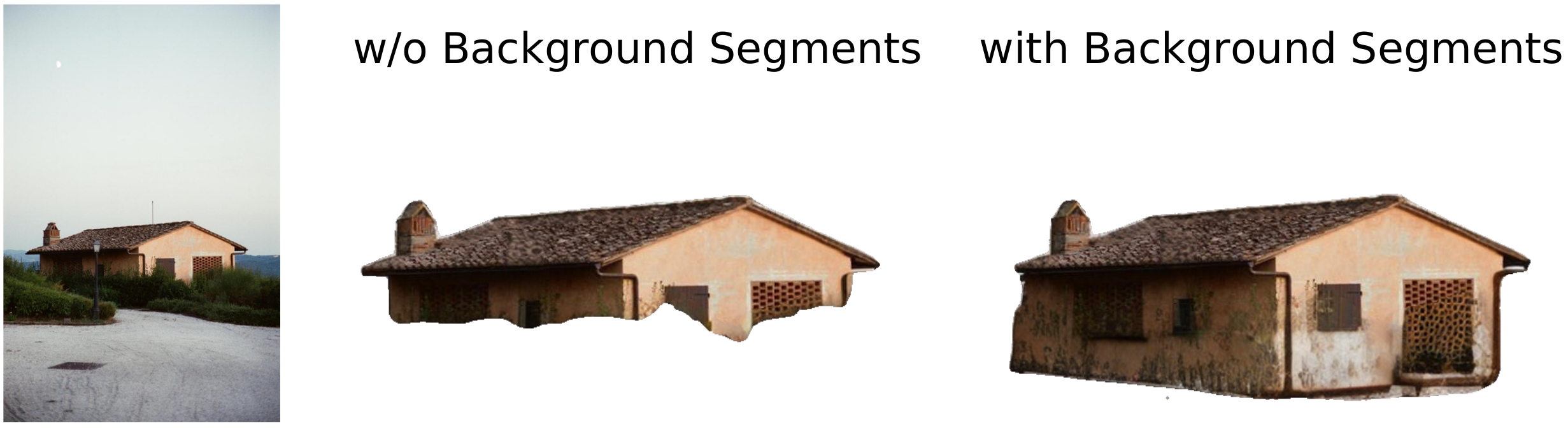}
   \caption{Amodal completion results from our method, with and without considering background segments as potential occluders.}
   \label{fig:background}
   \vspace*{-0.5cm}
\end{figure}

\textbf{Background Segments Effectiveness.}  Background segments (see~\cref{sec:textquery_segmentation}) contribute greatly to improving amodal completion in open-world contexts by handling occlusions and ambiguous backgrounds more effectively. As shown in~\cref{fig:background}, including background segments in the occlusion reasoning step makes it possible to capture the occlusion relationships between a query object and ambiguous background elements like foliage and ground, making scene understanding more comprehensive. This is particularly effective in the open-world setting, where occlusions are more diverse and complex.


\section{Conclusion}
\label{sec:conclude}
We presented a novel framework for open-world amodal appearance completion that reconstructs occluded objects across diverse categories without additional training. Our approach outperforms existing methods in user preference and objective quality metrics over an evaluation dataset of complex object occlusion cases, collected from four different data sources. This amodal completion framework readily supports applications like image editing, novel view synthesis, and 3D reconstruction~\cite{zhang2024transparent,tudosiu2024mulan,shi2023zero123++}, making it versatile for real-world use.

\textbf{Limitations.} Our reliance on pre-trained image generation models can occasionally introduce artifacts, such as mismatched poses in animal completions (e.g., generating a standing dog when it should be sitting based on the context; see appendix for more examples). However, our modular design allows for easy updates as better pre-trained models become available. Future work could improve inpainting accuracy and develop more comprehensive evaluation metrics, particularly for scenarios where ground-truth amodal data is unavailable.

\clearpage
\setcounter{page}{1}
\maketitlesupplementary

This supplementary material provides additional details, analyses, and resources to complement the main paper. Specifically, we include additional visual comparisons, showcasing more examples that compare our method with existing approaches in~\cref{sec:more_visual}. We then provide a failure analysis in~\cref{sec:failure_analysis}, categorizing failures into two types: complete failures, where no amodal completion is generated, with proportions of such failures for our method and comparison methods; and unsatisfactory completions, where the generated amodal result deviates from the expected appearance. In~\cref{sec:human}, the human study details are expanded upon with sample questionnaires used in the evaluations and inter-annotator agreement metrics, such as Fleiss' kappa, to establish the reliability of subjective assessments. The dataset collection process is described in detail in~\cref{sec:dataset_collect}, outlining the steps taken to create our evaluation dataset and including visualizations to demonstrate dataset diversity. Lastly, comprehensive configuration details are provided in~\cref{sec:configuration}, including the specifications of pre-trained models used, to support reproducibility and facilitate further research.

\section{Additional Visual Comparisons}
\label{sec:more_visual}

\begin{figure*}[!htbp]
  \centering
   \includegraphics[width=0.87\linewidth]{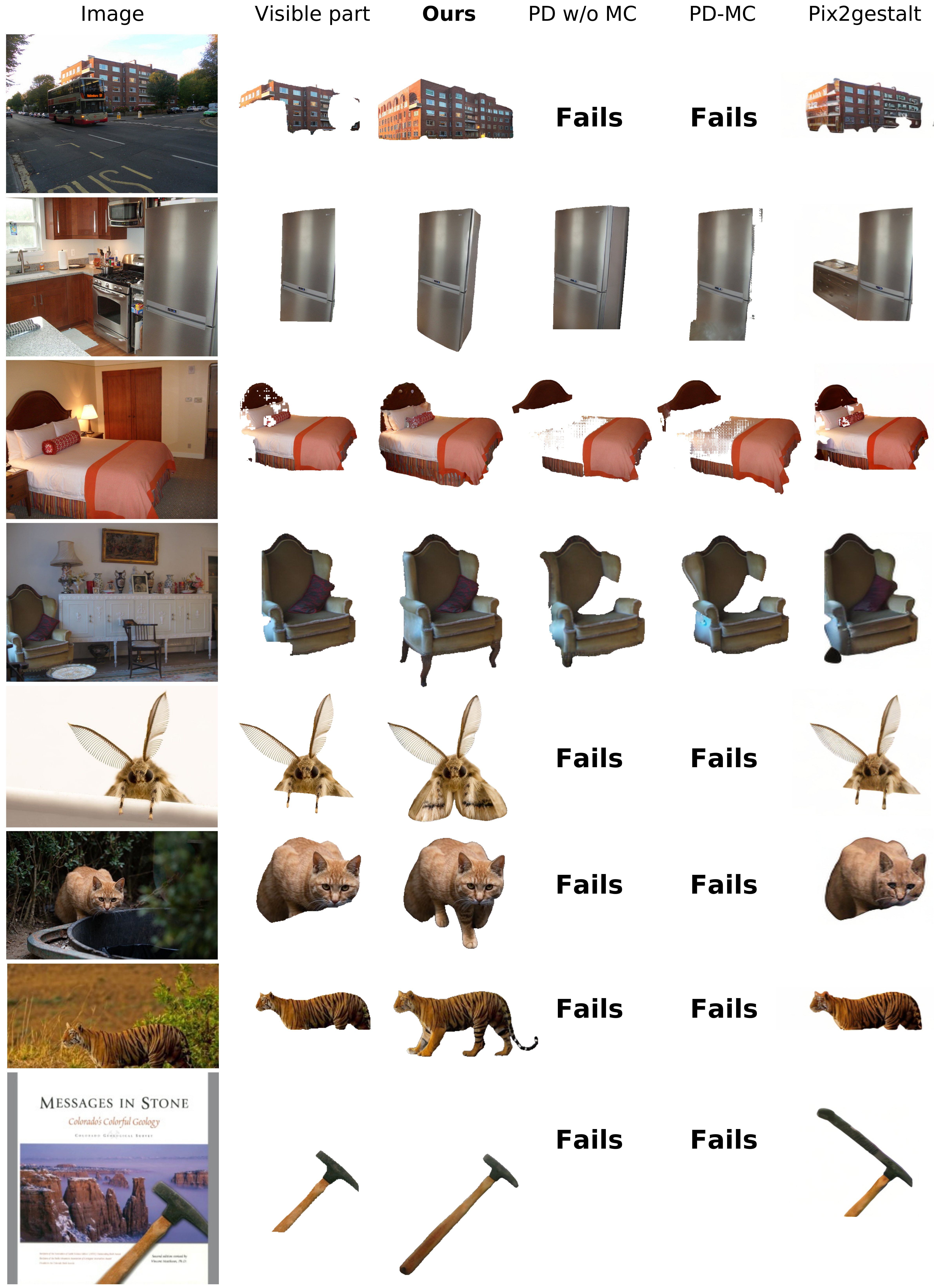}
   \caption{Visual comparisons of amodal completions across different methods. Our method consistently outperforms others in reconstruction integrity, handling complex occlusions, and producing plausible completions. Examples are drawn from VG, COCO-A, free images, and LAION datasets, with two images from each source. “Fails" refers in cases where no amodal completion result is generated.} 
   \label{fig:more_compare}
\end{figure*}

The visual comparisons in~\cref{fig:more_compare} showcase the effectiveness of our method in handling diverse and complex scenarios across a wide range of object categories and occlusion types. The examples span indoor, outdoor, and natural scenes, demonstrating our approach's open-world adaptability and robustness in producing realistic and complete amodal completions.

Our approach excels in reconstructing objects from diverse categories, including buildings, furniture, animals, insects, and tools. This adaptability underscores its ability to handle open-world scenarios without predefined object categories, unlike competing methods such as PD w/o MC and PD-MC~\cite{xu2024amodal}, which depend on fixed classes and fail when encountering unseen categories. In contrast, Pix2gestalt~\cite{ozguroglu2024pix2gestalt}, while always producing amodal completions due to its reliance on supervised learning with large training datasets, sometimes minimally alters the input image (e.g., the moth and cat examples), offering little meaningful reconstruction.

\begin{figure}[!htbp]
  \centering
   \includegraphics[width=\linewidth]{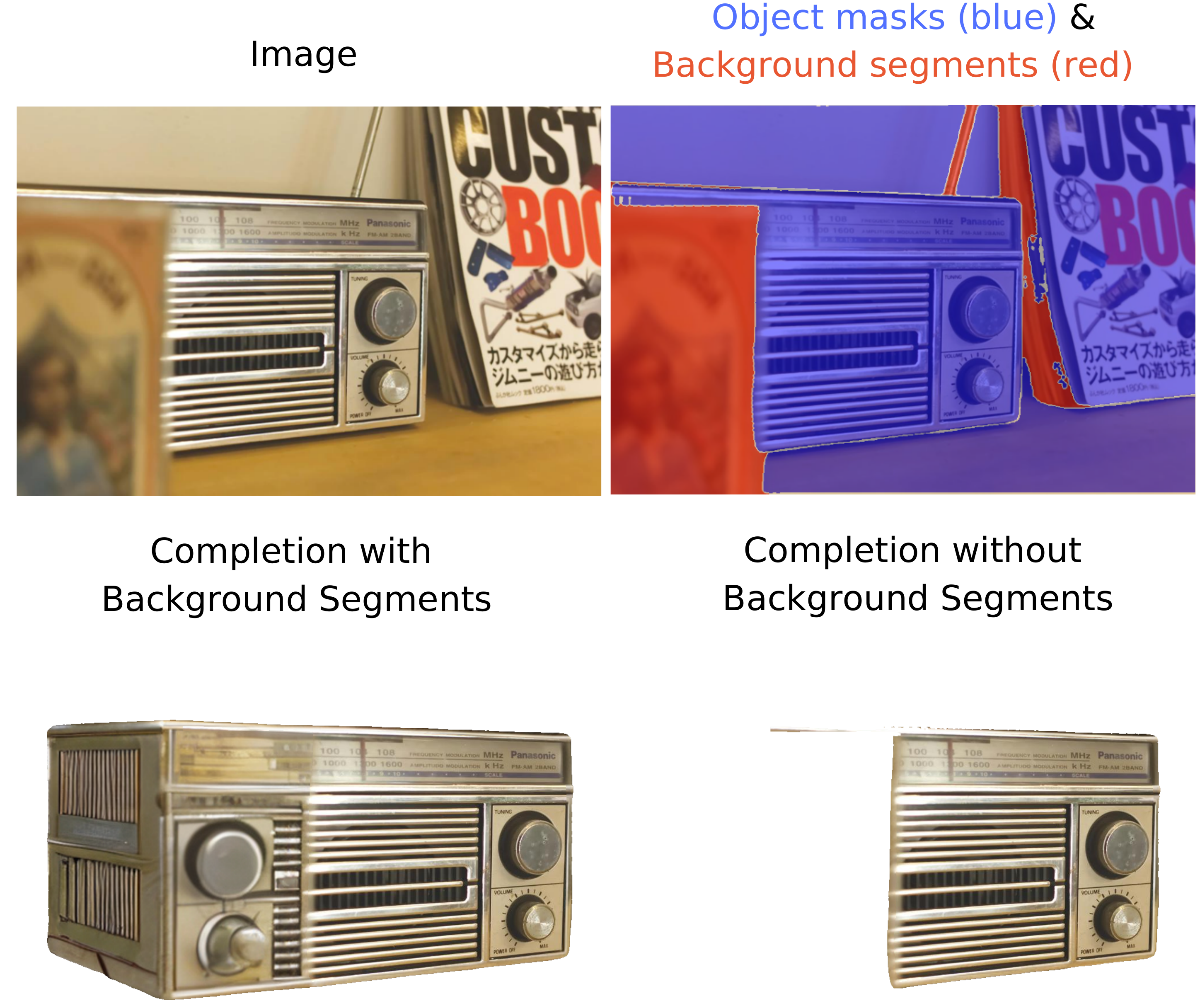}
   \caption{Comparison of our amodal completion results with and without considering background segments. Object masks are shown in blue, and background segments are highlighted in red. Our method with background segmentation reconstructs the occluded object more comprehensively, capturing missing structural details (e.g., the side panel of the radio) that are ignored when background segments are not utilized.}
   \label{fig:segments}
\end{figure}

Our method demonstrates the ability to handle complex occlusions, such as objects occluded by shrub or fences, mutual overlapping elements, or ambiguous background regions. Examples such as the building, moth, and cat emphasize the critical role of our background segmentation strategy in identifying and effectively handling occluding background segments. This strategy improves structural integrity in scenarios where competing methods fail entirely (e.g., PD w/o MC and PD-MC) or produce incomplete reconstructions (e.g., Pix2gestalt).

\textbf{Impact of Background Segmentation.} \cref{fig:segments} demonstrates the importance of incorporating background segments into our method. Our background segmentation strategy successfully identifies and accounts for ambiguous background regions, such as the red-highlighted areas. With background segments considered, the amodal completion preserves structural integrity and reconstructs missing details, such as the side panel of the radio.

\section{Failure Analysis} 
\label{sec:failure_analysis}

We analyse failure cases in two types: (i) complete failures, where no amodal completion is generated, and (ii) partial failures, where the generated completion is unsatisfactory. 

\begin{figure*}[!t]
  \centering
   \includegraphics[width=0.87\linewidth]{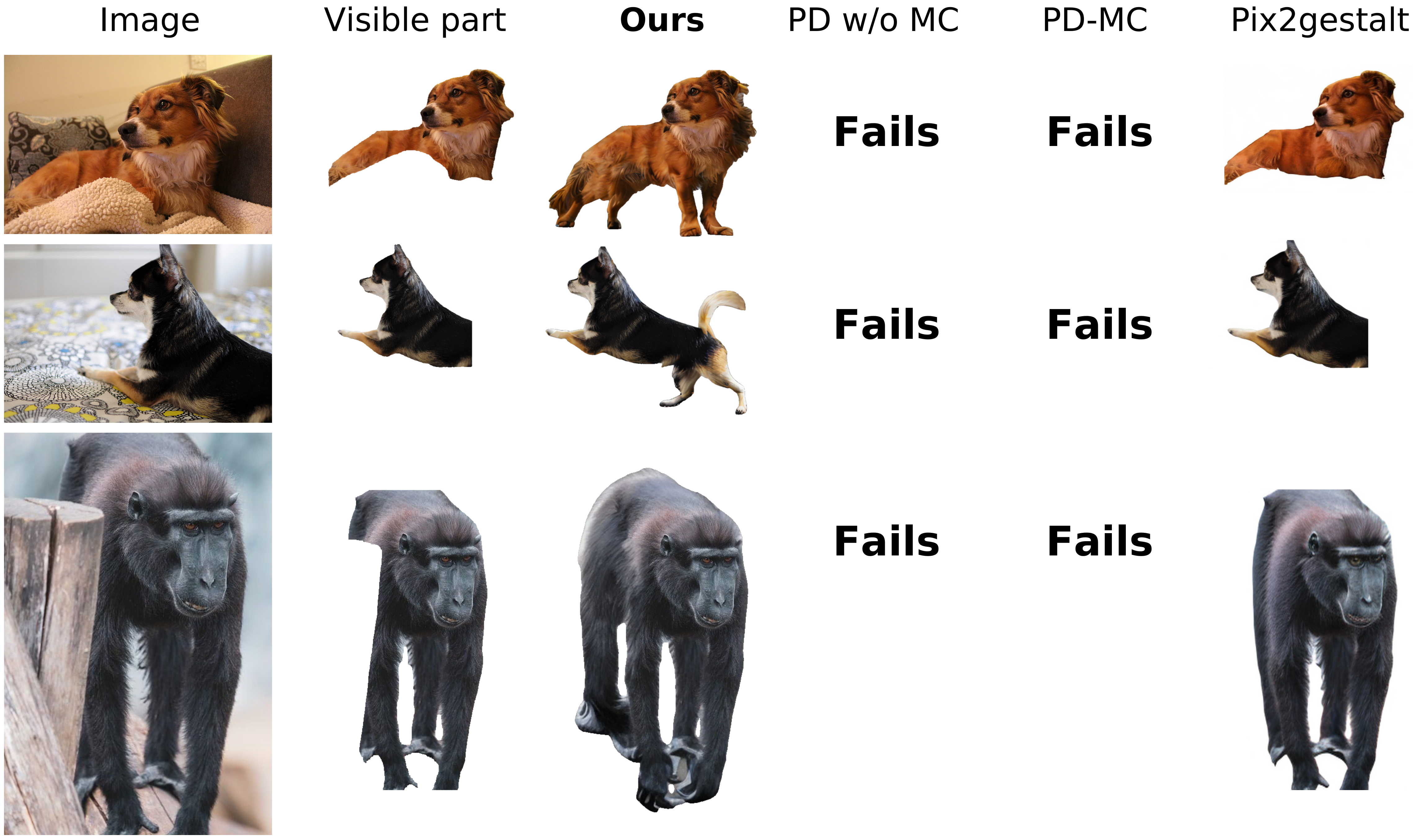}
   \caption{Examples of our common partial failures, such as unnatural poses and hand reconstruction. “Fails" refers in cases where no amodal completion result is generated. Other methods have either completely failed or partially failed in these cases. Compared to other methods, our framework maintains structural integrity, even if some results are not satisfactory.}
   \label{fig:partial_failures}
\end{figure*}

\textbf{Complete Failures.} \cref{tab:fails_rate} details the proportion of complete failures—cases where no amodal completion result is generated—for each method across datasets. While Pix2gestalt achieves a 0\% failure rate due to its supervised design, it sometimes minimally alters the input without meaningfully addressing occlusion, as seen in~\cref{fig:more_compare} (e.g., moth and cat examples). In contrast, our method maintains a low failure rate (4.1\% overall) while ensuring structural integrity and realistic completions. Unlike PD w/o MC and PD-MC, which exhibit overall failure rates exceeding 40\% across datasets, our approach succeeds in reconstructing occluded objects even under challenging conditions. Our failures occur primarily in cases where occluded objects are undetectable. Notably, our method avoids meaningless pixel modifications by not attempting amodal completions when occluded objects are undetectable.

\begin{table}[!t]
\setlength{\tabcolsep}{1pt}
\centering
\footnotesize
\begin{tabular}{cc|ccccc}
\toprule
            & \begin{tabular}[c]{@{}c@{}}Training \\ Free\end{tabular} & VG~\cite{krishna2017visual}    & COCO-A~\cite{zhu2017semantic} & \begin{tabular}[c]{@{}c@{}}Free \\ Image\end{tabular} & LAION~\cite{schuhmann2021laion} & Overall \\ \hline
PD w/o MC   & Yes           & 39.6\% & 47.3\%  & 58.2\%      & 53.1\% & 44.9\%   \\
PD-MC       & Yes           & 40.0\% & 48.5\%  & 58.2\%      & 54.2\% & 45.5\%   \\
Pix2gestalt & No            & 0.0\%  & 0.0\%   & 0.0\%       & 0.0\%  & 0.0\%    \\
Ours        & Yes           & 6.3\%  & 1.8\%   & 2.0\%       & 1.7\%  & 4.1\%    \\ \bottomrule
\end{tabular}
\caption{Proportion of complete failures (no amodal completion generated) for each method across datasets.}
\label{tab:fails_rate}
\end{table}

\textbf{Partial Failures.} While our method excels in most cases, occasionally partial failures occur, where the generated completion does not match expectations. \cref{fig:partial_failures} showcases common partial failures, such as generating a standing dog instead of a sitting one based on the context, or producing gorilla hands with unnatural shapes. These issues arise primarily due to limitations in pre-trained models used for inpainting and the inherent ambiguity in certain occlusions. Unlike other methods, however, our framework maintains structural integrity, even in partially unsatisfactory results. Future work could address these limitations by refining inpainting models and incorporating additional contextual reasoning mechanisms.

\section{Human Study Design and Inter-Participant Agreement}
\label{sec:human}
To evaluate the subjective quality of amodal completions, we conducted a structured human study designed to assess the realism and completeness of generated results across various methods. Using the Prolific crowdsourcing platform\footnote{\url{https://www.prolific.com/}}, we recruited 180 participants to compare our method against Pix2gestalt~\cite{ozguroglu2024pix2gestalt}, PD w/o MC, and PD-MC~\cite{xu2024amodal}. Each participant was presented with an original image alongside four completed versions corresponding to the methods under evaluation. The order of the four completions was randomized to mitigate positional bias. Participants were tasked with selecting the completion that best represented a realistic and whole view of the object, based on visible cues from the original image. 

\begin{figure}[!t]
  \centering
   \includegraphics[width=0.95\linewidth]{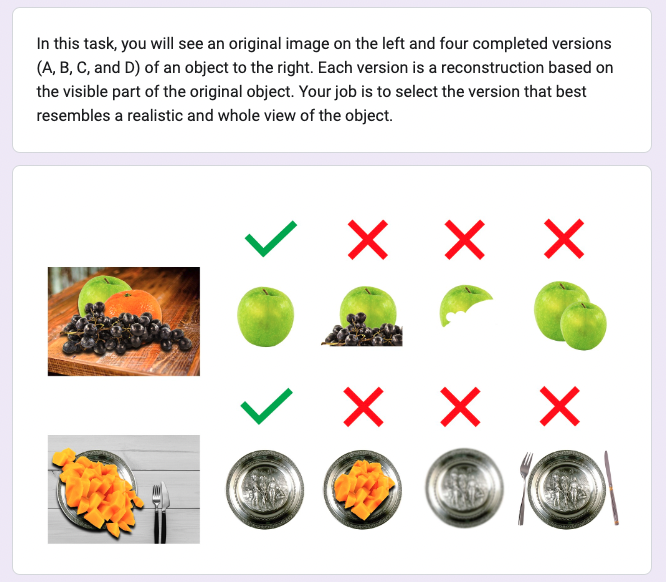}
   \caption{A detailed guide provided to participants at the beginning of the questionnaire, demonstrating how to assess the realism and completeness of amodal completions}
   \label{fig:questionnaire_guide}
\end{figure}

\begin{figure}
  \centering
   \includegraphics[width=0.95\linewidth]{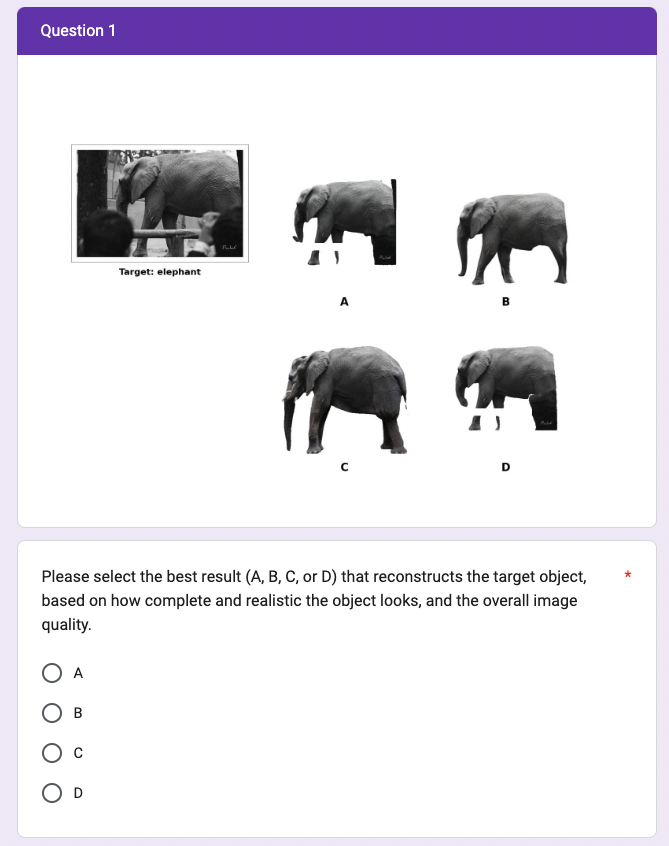}
   \caption{An example question from the human study, showing the original image (left) and four completions in randomized order (right). Participants were instructed to select the version that best reconstructed the target object.}
   \label{fig:questionnaire_sample}
\end{figure}

To ensure the task's clarity, we provided participants with a detailed guide outlining the evaluation process (see~\cref{fig:questionnaire_guide}). Additionally, \cref{fig:questionnaire_sample} illustrates a sample question interface. To ensure data reliability, 10\% of the questions were “gold standard” trials with unambiguous correct answers. Only participants who passed at least 75\% of these quality control checks were included in the final analysis. The “gold standard” trials were not included in the main data analysis.

\textbf{Inter-Participant Agreement.} To quantify the inter-participant agreement across multiple raters, we computed Fleiss' kappa scores for each dataset and the overall study. \cref{tab:fleiss} summarizes the calculated $\kappa$ values, which consistently fall within the “fair agreement" range~\cite{landis1977measurement}. Among the datasets, the highest agreement was observed for the LAION ($\kappa$ = 0.374), while the overall agreement for the study was $\kappa$ = 0.319. This indicates a fair level of consistency among participants despite the inherent subjectivity of the task.

\begin{table}[!h]
\setlength{\tabcolsep}{4pt}
\centering
\begin{tabular}{ccccc}
\toprule
VG~\cite{krishna2017visual}    & COCO-A~\cite{zhu2017semantic} & \begin{tabular}[c]{@{}c@{}}Free\\ Image\end{tabular} & LAION~\cite{schuhmann2021laion} & Overall \\ \hline
0.275 & 0.364 & 0.353                                                 & 0.374 & 0.319   \\ \bottomrule
\end{tabular}
\caption{Fleiss' kappa between human participants.}
\label{tab:fleiss}
\end{table}

The observed agreement reflects the complexities involved in assessing amodal appearance completions, which require judgments on both perceptual realism and context-dependent plausibility. While variability in individual preferences is expected, the fair consistency across all datasets highlights the reliability of our evaluation framework. This findings also shows the importance of subjective evaluation in capturing perceptual nuances that quantitative metrics may overlook.

\section{Dataset Collection}
\label{sec:dataset_collect}

Our evaluation dataset integrates images from four sources: COCO-A~\cite{zhu2017semantic}, Visual Genome (VG)~\cite{krishna2017visual}, LAION~\cite{schuhmann2021laion}, and copyright-free images collected from publicly accessible websites\footnote{\url{https://www.pexels.com/}}\footnote{\url{https://pixabay.com/}}\footnote{\url{https://unsplash.com/}}. A total of three human annotators collected images containing occluded objects from different sources, and each independently provided a category label for one or more occluded objects in the image. The resulting dataset consists of 2379 images spanning 553 distinct target object classes.

\begin{figure}[!t]
  \centering
   \includegraphics[width=\linewidth]{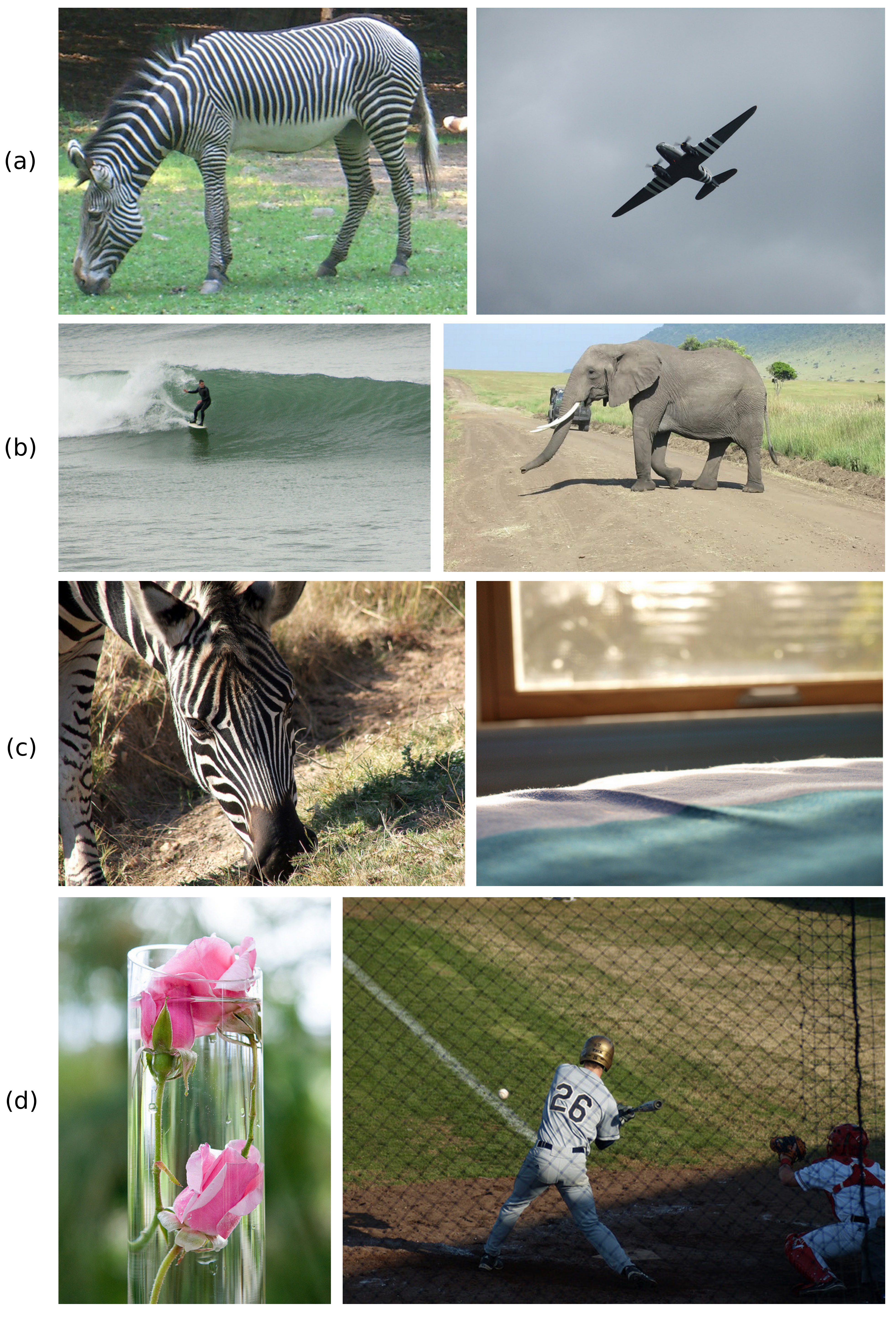}
   \caption{Examples of filtered images from COCO-A~\cite{zhu2017semantic}: (a) Background elements occluded but primary objects visible (e.g., zebra, airplane). (b) Visible object area below 2\% of the total image area (e.g., surfboard, vehicle). (c) Most of the primary object lies outside the image boundary. (d) Occluders are transparent or excessively thin (e.g., glass, wires).}
   \label{fig:filter}
\end{figure}

COCO-A contribute natural scenes with realistic occlusions, providing a foundation of everyday scenarios and common objects. However, not all images in COCO-A feature object-specific occlusions since it was originally designed for semantic segmentation~\cite{zhu2017semantic}. To address this, we applied a filtering process that removed images where (a) background elements were occluded but primary objects were not, (b) the visible part of the primary object occupied less than 2\% of the total image area, (c) most of the primary object lay outside the image boundary, or (d) occluders were transparent or excessively thin (e.g., glass or wires). \cref{fig:filter} illustrates examples of filtered images from COCO-A.

\begin{figure}[!t]
  \centering
   \includegraphics[width=\linewidth]{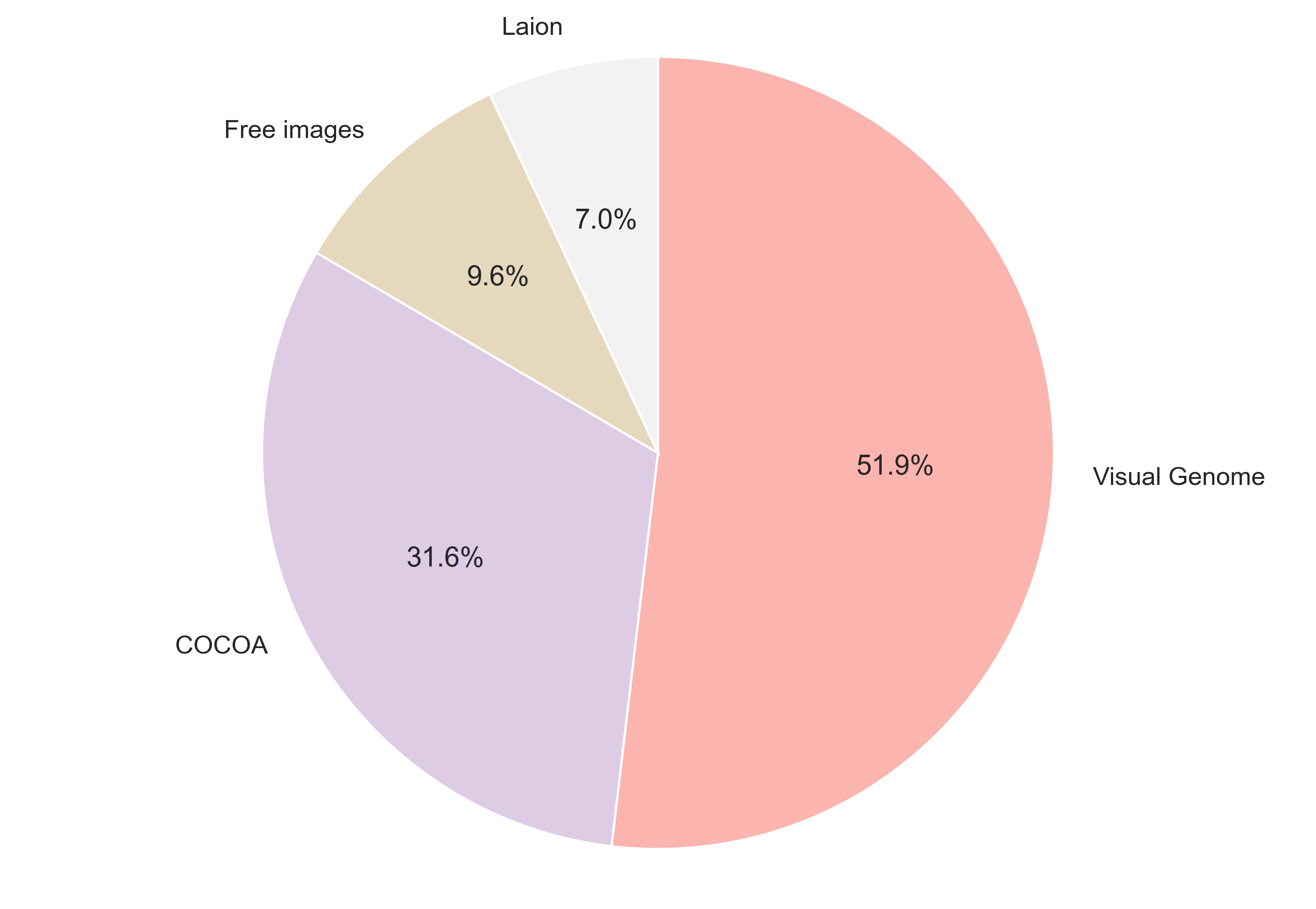}
   \caption{Image distribution across different datasets.}
   \label{fig:image_distribution}
\end{figure}

To further enhance diversity, we incorporated images from VG~\cite{krishna2017visual}, LAION~\cite{schuhmann2021laion} and copyright-free sources, introducing a broad range of lighting conditions, object appearances, and complex occlusions typical of unconstrained, open-world environments. \cref{fig:image_distribution} shows the distribution of images across the four dataset sources. VG accounts for the largest share at 51.9\%, followed by COCO-A (31.6\%), copyright-free images (9.6\%), and LAION (7.0\%). The combination of these sources ensures that our evaluation dataset captures a wide range of real-world occlusion scenarios and object categories.

\begin{figure}[]
  \centering
   \includegraphics[width=\linewidth]{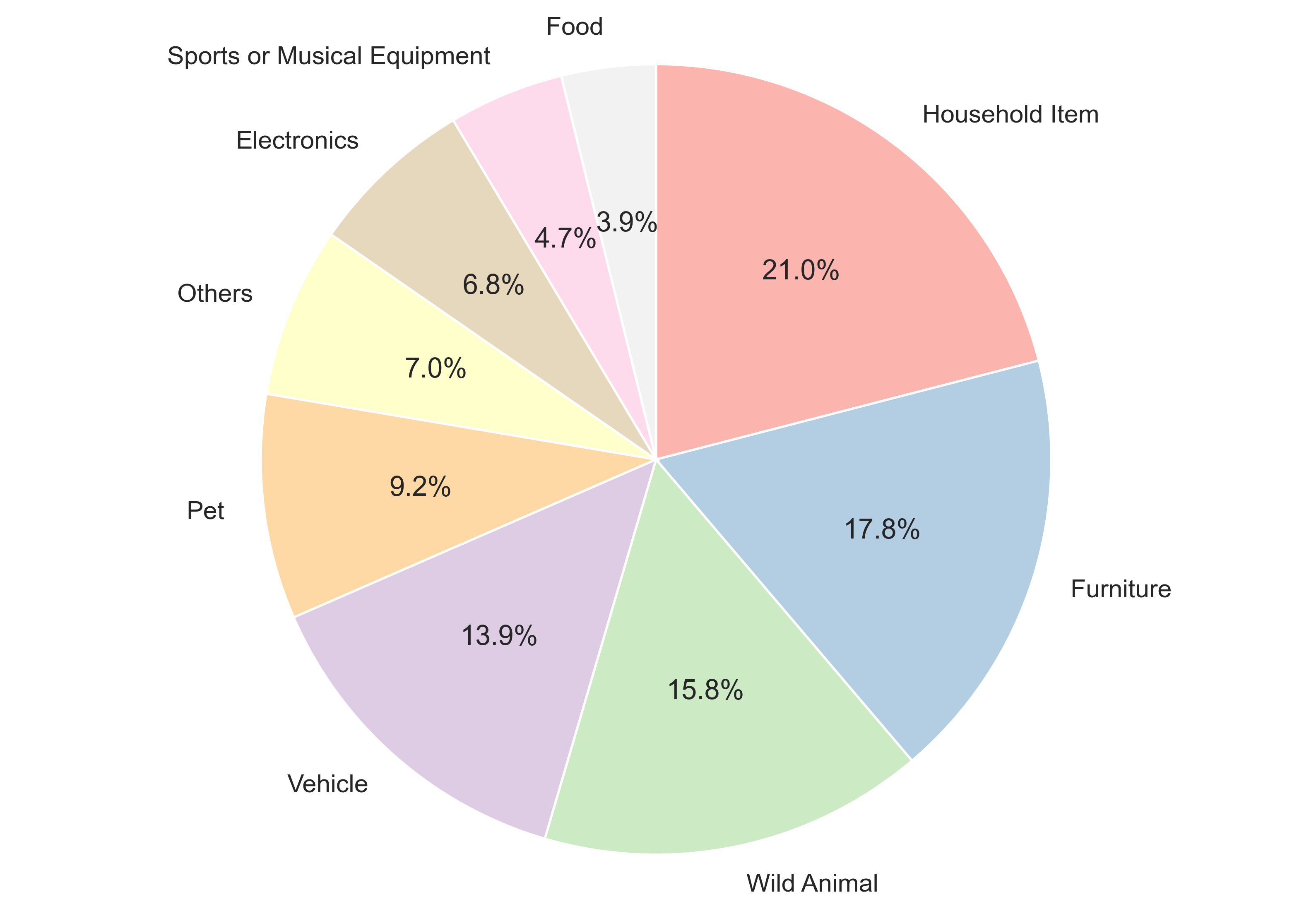}
   \caption{Distribution of object categories in major groups.}
   \label{fig:image_group}
\end{figure}

To analyse our evaluation dataset composition, we further grouped object labels into broad categories based on their semantic meanings. This grouping provides insights into the types of objects present in the dataset. As shown in~\cref{fig:image_group}, prominent categories include “Household Item" (21\%), “Furniture" (17.8\%), and “Wild Animal" (15.8\%), reflecting the dataset's relevance to both everyday scenarios and naturalistic environments. Other categories, such as “Vehicle", “Pet", and “Food" further ensure coverage of real-world contexts across various settings. 

The grouping of object labels into these broad categories was based on specific mapping rules. For instance, “Household Item" encompasses items commonly found in daily home life, including objects like bowls, spoon and scissors. The “Furniture" category includes various types of household furnishings such as sofas, chairs and tables. Domesticated animals such as dogs and cats were categorized under “Pet", while “Wild Animal" includes non-domesticated species like elephants, giraffes and bear. Transportation-related objects, such as trucks, bicycles, and airplanes, were grouped into the “Vehicle" category, whereas electronic devices like laptops, phones and cameras were placed in “Electronics". The “Sports or Musical Equipment" category includes items like tennis rackets, guitars, and piano, covering recreational and artistic tools. “Food" represents a variety of edible items, including apples, bread, and pizza. Finally, a catch-all “Others" category includes objects that do not fit neatly into the previous groups, such as buildings and natural elements (e.g., trees, flowers).

\section{Configuration Details}
\label{sec:configuration}

To ensure reproducibility and facilitate further research, this section details the configuration and pre-trained models used in our framework. 

For vision language-grounded object identification, we utilized LISA-13B-llama2-v1 model~\cite{lai2024lisa}, which offers robust reasoning capabilities for mapping natural language queries to visible object regions. The appearance of the occluded regions were reconstructed using the Stable Diffusion v2 inpainting model~\cite{rombach2022high}, known for its high-quality generative performance. To enhance scene understanding and support object detection, we incorporated the RAM++ image tagging model (ram\_plus\_swin\_large\_14m)~\cite{huang2023open}, enabling open-set tagging of visual elements, and the GroundingDINO object detector (groundingdino\_swint\_ogc)~\cite{liu2024grounding}, which effectively identifies and segments objects in open-world settings.

To assess occlusion relationships, we employed InstaOrderNet (InstaOrder\_InstaOrderNet\_od)~\cite{lee2022instance}, a model pre-trained for amodal occlusion ordering tasks. This model processes pairwise object masks and image patches without relying on object category labels, making it suitable for the diverse and ambiguous occlusions in open world scenes. For pixel-wise segmentation tasks, we used the Segment Anything model (sam\_vit\_h\_4b8939)~\cite{kirillov2023segment}, which provided accurate segmentation across various object and background types. The CLIP model (ViT-B/32)~\cite{radford2021learning} was utilized for text-image alignment during inpainting prompt generation, leveraging its powerful feature extraction for matching visual and semantic cues.

These pre-trained models, each specializing in a specific subtask, enabled us to construct a robust framework tailored to the challenges of open-world amodal appearance completion. Leveraging their embedded knowledge allowed us to address complex, real-world scenarios without requiring additional training. Furthermore, the modularity of our framework ensures that each component can be easily replaced with improved pre-trained models as they become available, enhancing adaptability and future extensibility.

{
    \small
    \bibliographystyle{ieeenat_fullname}
    \bibliography{main}
}

\end{document}